\documentclass[wcp]{jmlr}

\usepackage{longtable}
\usepackage{amsmath,amssymb,amsfonts}
\DeclareMathOperator*{\argmax}{arg\,max}
\usepackage{graphicx}
\usepackage{textcomp}
\usepackage{algorithm}
\usepackage{algorithmicx}
\usepackage{pifont}
\usepackage{algpseudocode}
\usepackage{booktabs}

\makeatletter
\def\set@curr@file#1{\def\@curr@file{#1}} 
\makeatother
\usepackage[load-configurations=version-1]{siunitx} 

\jmlrvolume{129} 
\editors{Sinno Jialin Pan and Masashi Sugiyama}
\jmlryear{2020}
\jmlrworkshop{ACML 2020}

\title[Short Title]{Atlas-aware ConvNet \\ for Accurate yet Robust Anatomical Segmentation}

  \author{\Name{Yuan Liang} \Email{liangyuandg@g.ucla.edu}\\
   \Name{Weinan Song} \Email{wsong@ucla.edu}\\
   \Name{Jiawei Yang} \Email{jiawei118@g.ucla.edu}\\
   \Name{Liang Qiu} \Email{liangqiu@g.ucla.edu}\\
   \Name{Kun Wang} \Email{wangk@ucla.edu}\\
    \Name{Lei He} \Email{lhe@ee.ucla.edu}\\
   \addr University of California, Los Angeles, CA 90095, USA}

\begin{document}

\maketitle

\begin{abstract}
Convolutional networks (ConvNets) have achieved promising accuracy for various anatomical segmentation tasks. 
Despite the success, these methods can be sensitive to data appearance variations. 
Considering the large variability of scans caused by artifacts, pathologies, and scanning setups, robust ConvNets are vital for clinical applications, while have not been fully explored.
In this paper, we propose to mitigate the challenge by enabling ConvNets' awareness of the underlying anatomical invariances among imaging scans.
Specifically, we introduce a fully convolutional Constraint Adoption Module (CAM) that incorporates probabilistic atlas priors as explicit constraints for predictions over a locally connected Conditional Random Field (CFR), which effectively reinforces the anatomical consistency of the labeling outputs.
We design the CAM to be flexible for boosting various ConvNet, and compact for co-optimizing with ConvNets for fusion parameters that leads to the optimal performance.
We show the advantage of such atlas priors fusion is two-fold with two brain parcellation tasks.
First, our models achieve state-of-the-art accuracy among ConvNet-based methods on both datasets, by significantly reducing structural abnormalities of predictions. 
Second, we can largely boost the robustness of existing ConvNets, proved by: (i) testing on scans with synthetic pathologies, and (ii) training and evaluation on scans of different scanning setups across datasets.
Our method is proposing to be easily adopted to existing ConvNets by fine-tuning with CAM plugged in for accuracy and robustness boosts. 
\end{abstract}
\begin{keywords}
Anatomical segmentation, machine learning, convolutional neural networks. 
\end{keywords}

\section{Introduction}
Anatomical segmentation performs the annotation of tissues and structures for medical images. 
Currently, ConvNets have achieved promising accuracy for this task (\cite{ronneberger2015u,li2017compactness,milletari2016v,liang2019comparenet}), typically by employing stacked convolutional layers to learn hierarchical features for voxel-wise classification. 
Despite the success, achieving robust segmentation with ConvNets is still challenging: 
the ConvNets cannot easily generalize to scans with unforeseen appearance variations, which might be caused by artifacts (\cite{he2019non}), pathologies (\cite{dey2018compnet}), and different acquisition procedures across scanners (\cite{leung2010robust}).
This challenge of robustness is even more significant since the medical datasets are mostly compact, due to the labeling cost and privacy issue, such that ConvNets cannot efficiently learn generalized features. 

One promising approach to the challenge is by incorporating ConvNets with prior knowledge about anatomies. 
Specifically, similar anatomical structures, \textit{e.g.} anatomy shapes and locations, are shared across scans of different individuals, and fusing such information with ConvNets can potentially facilitate the segmentation.
Several existing works have explored this approach, and they mostly fall into two categories: loss-based methods and graph-based methods.

\paragraph{Loss-based methods} perform knowledge fusion by constraining ConvNet predictions with pre-defined statistics about anatomies as parts of training objectives. 
For example, loss terms for regularizing anatomical adjacency (\cite{ganaye2018semi,bentaieb2016topology}) and boundary conditions (\cite{chen2017dcan}) have been applied to reduce
prediction abnormalities. 
Distances between predictions and empirical distributions of organ sizes (\cite{zhou2019prior,kervadec2019constrained}) and latent features (\cite{oktay2017anatomically,dalca2018anatomical}) have also been exploited for optimization. 
However, such methods only utilize parts of anatomical invariances that either distilled from training or handcrafted as metrics. 
Moreover, they mostly require designing task-specific losses that cannot directly adopted to solve general anatomical segmentations. 

\paragraph{Graph-based Methods} construct graphical models that containing both atlas priors, \textit{i.e.} probability maps that spatially encode chances of observing each anatomy, and a likelihood from image appearances. 
Different from setting training objectives, the prior term can better capture high-resolution anatomical knowledge.
Classical atlas-based methods (\cite{patenaude2011bayesian,sabuncu2010generative,fischl2002whole}) have widely applied this approach, mostly employing appearance model of simple forms (\cite{iglesias2015multi}) to estimate anatomical probabilities from image intensities. 
One work (\cite{gao2016segmentation}) improves the accuracy by utilizing a high discriminative ConvNet to characterize the
appearance-based probability. 
However, under the existing heavy graph designs, parameters of the prior model and those of the ConvNet-based appearance model can only be estimated separately, making the prior fusion as a post-processing step for ConvNets predictions.

In this work, we introduce a novel graphical model that enables the optimal fusion of atlas priors and ConvNets with the end-to-end training.
The core of the method is a Constraint Adoption Module (CAM) which formulates a Conditional Random Field (CRF) over the target scan and a prior atlas. 
The CRF models atlas priors as soft constraints for the predictions of ConvNets, and is locally connected to retain the spatial structure of anatomical invariances. 
We also show the connection locality enables the
CAM to be fully convolutional and compact, such that it can
be optimized with any ConvNet by back-propagation without
permutohedral lattice approximations.

Our method is distinct from existing works (\cite{zheng2015conditional,arnab2016higher,zhao2016brain,fu2016deepvessel}) that formulate CRFs inference as Recurrent Neural Networks for end-to-end segmentation, mainly from two aspects: (\textit{i}) our graph is over two volumes, a target scan and an atlas, to enable the label propagation for anatomical consistency of the labeling, while theirs are within the target image for smooth predictions; (\textit{ii}) they mostly employ graphs of full connectivity for discrimination, which is computational costly for 3D data and unreasonable for medical scans. 
We have performed experiments on two brain benchmarks (MICCAI and IBSRv2) with three ConvNets of different architectures as backbone. 
Results show our method significantly increases the accuracy (increasing Dice Score of up to 4.38\%) by reducing anatomical inconsistencies in predictions.
Meanwhile, our method boosts the robustness of ConvNets, verified by testing on scans with unforeseen pathologies (reducing Local Dice Decay to up to 0.46$\times$), and by cross-dataset evaluation with scans of different distributions (reducing Dice Score changes to up to 0.40$\times$). 

In general, our contributions can be summarized as follows:
\begin{itemize}
  \item We present the first end-to-end graphical method to incorporate atlas priors to ConvNets for anatomical segmentation. We demonstrate the method is general and flexible to boost various ConvNets. 
  \item We propose a locally connected graph design over two volumes for label propagation, and formulate its computation to be fully convolutional with compactness. 
  \item We verify the accuracy and robustness boosting effect of our method, and perform ablation tests to investigate the impact of different model potentials to the performance. 
\end{itemize}

\section{Methodologies}
\begin{figure}[ht]
\centering
\includegraphics[width=\textwidth]{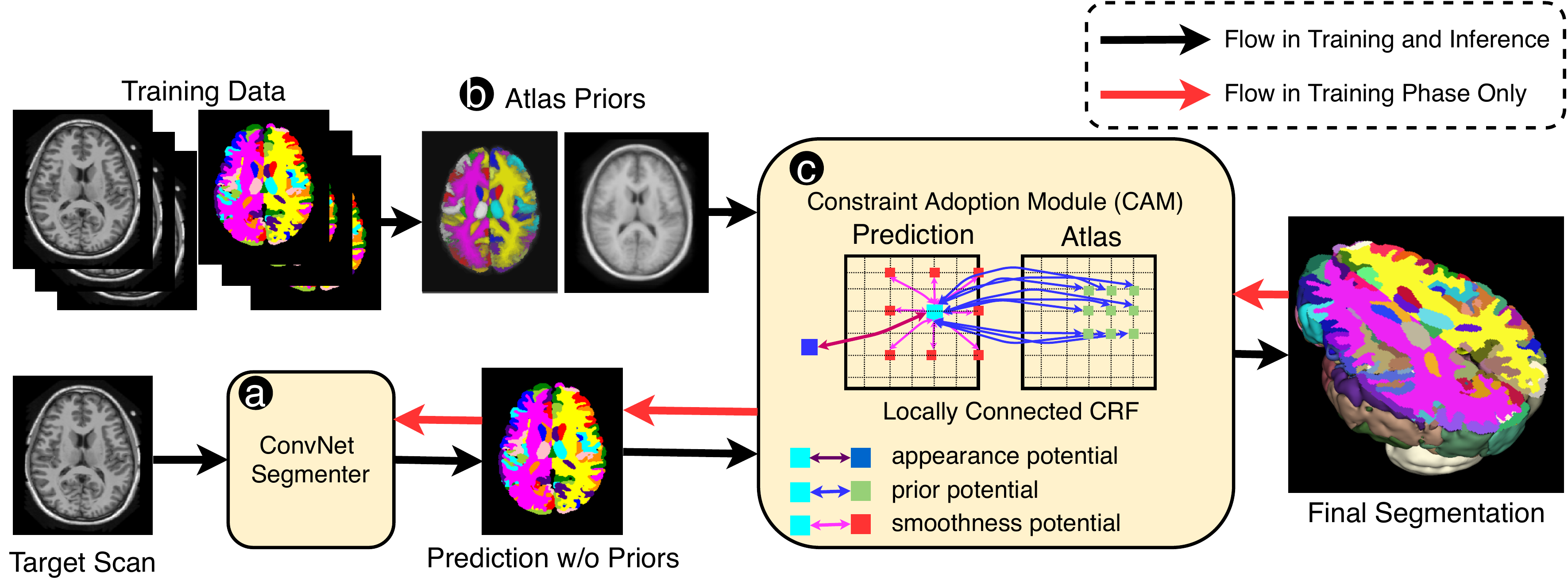}
\caption{Overall architecture of our method. 
An atlas pair of scan and its probabilistic labeling map is pre-generated from training data to encode atlas priors. 
A Constraint Adoption Module (CAM) constructs a locally-connected CRF to fuse ConvNet-based appearance potentials, prior potentials and smoothness potential for anatomical consistent segmentation. 
CAM is fully convolutional and compact for end-to-end training by formulating its mean field inference as convolutions without permutohedral lattice approximations. }
\label{arth}
\end{figure}

Figure \ref{arth} demonstrates the overall architecture of our proposed method. 
An atlas pair, which consists of an atlas scan and a probabilistic atlas label that generated from training data, is adopted with the Constraint Adoption Module (CAM) as constraints of predictions. 
CAM formulates a locally connected CRF over the prediction and
atlas, which models: (i) appearance potentials characterized
by a ConvNet segmenter, (ii) prior potentials for propagating
atlas priors into label fusion, and (iii) smoothness potentials for encouraging local appearance consistency within predictions.
The whole model is fully convolutional and compact for training with back-propagation. 
We elaborate the details of ConvNet segmenters, probabilistic atlas pair, CAM, and training strategies in the next sections.

\subsection{ConvNet Segmenter}
Our method relies on a ConvNet (Fig. \ref{arth}(a)) to produce an appearance-based prediction of anatomies given image patches.
Note the segmenter can be any ConvNet architecture
that is end-to-end trainable, and in this work, we adopt three popular networks of different architectures as the segmenter to illustrate the flexibility of our method: UNet (\cite{ronneberger2015u}), VNet (\cite{milletari2016v}) and HCNet (\cite{li2017compactness}). 
Both UNet and VNet utilize the encoder-decoder structure, while differences exist in feature map encoding, feature forwarding, and model configuration. 
Meanwhile, HCNet utilizes dilated convolutions, where the feature map keeps at the high resolution without encoding or decoding. 
Since the CAM is implemented to be differentiable, we optimize the parameters of ConvNet segmenters together with the CAM.

\subsection{Atlas Prior Generation}
To construct a representative atlas pair, we first normalize all the training data, and finely align them to the standard MNI-152 template (\cite{fonov2011unbiased}) with syn registrations (\cite{avants2008symmetric}).
Then, we define the atlas scan as the average intensity map of the co-registered scans, and the atlas label as the average probability map of observing each anatomy type. 
Note that in applications, any available scan and annotation can be applied to construct the atlas for possibly better prior generalisations, while in this work we only alleviate those from the public datasets.
Figure \ref{arth}(b) includes an example atlas pair
generated from MICCAI2012 dataset, where the opacity of the
label map reflects the probability of the dominant anatomies.
Although more sophisticated atlas construction methods exist, \textit{e.g.} \cite{blaiotta2018generative}, for capturing peculiar anatomical features, this simple method can already illustrate the effectiveness of CAM. 
During the forward propagation of the model, we align the
prior atlas pair to the target scan with simple affine registration to reduce their geometric gaps of anatomical structures.
The finer registrations are not applied since the connectivity of CAM tolerates local alignment deviations, and can avoid adding extra runtime during inference. 

\subsection{Constraint Adoption Module (CAM)}
Let $T$ be a target 3-D scan volume defined over a discrete domain $\Omega \subset \mathbb{R}^3$, and $S_{T}$ be its multi-class segmentation map. 
Let $\{A, S_{A}\}$ be the derived atlas prior pair, consisting of an atlas volume $A$ and its probabilistic label map $S_{A}$. 
CAM formulates the segmentation as the maximum-a-posteriori estimation of the label map $L_{T}$ conditioned on the target scan $T$ and the aligned probabilistic atlas pair $\{A, L_{A}\}$: 
\begin{equation}
    \hat{L_{T}} = \argmax_{l_{T}}P(l_{T}|X, A, L_{A}). 
\label{eq1}
\end{equation}
We consider $L_{T},L_{A}$ to be two random fields of labels that defined over sets of variables taking values from the labeling category set. 
We also consider $T,A$ to be two random fields of appearance that defined over sets of variables that represent voxel intensities. 
We characterize the CRF ($L_{T}, T, A, L_{A}$) with a Gibbs distribution $P(L_{T}=l_{T})=\frac{1}{Z}exp(-\sum_{c \in C}\phi_{c}(l_{T}|T, A, L_{A}))$, where $Z$ is a partition function, $C$ defines all the connections within the graph that induce Gibbs energy potentials, and $l_{T}$ is a labeling instance. 
As shown in Fig. \ref{arth}(c), we include three types of energy potentials to measure the cost for assigning the label $l_{T}$: (\textit{i}) appearance potentials $\phi_{a}(l_{T}|T)$ that conditioned on the volume appearance, (\textit{ii}) prior potentials $\phi_{p}(l_{T}|T, A, L_{A})$ that conditioned on the observation of atlas labels, and (\textit{iii}) smoothness potential $\phi_{s}(l_{T}|T)$ that conditioned on the labels of adjacent voxels. 
Formally, the conditional probability of labeling $l_{T}$ can be defined by: 
\begin{equation}
    \begin{array}{l}
        P(l_{T}|T,A,L_{A}) 
        = \frac{1}{Z}\exp(-\sum_{C_{a}}{\phi_{a}(l_{T}|T)}) \\
        ~~~~~~~~~~~~~~~-\sum_{C_{p}}\phi_{p}(l_{T}|T,A,L_{A}) -\sum_{C_{s}}\phi_{s}(l_{T}|T), 
    \end{array}
\label{eq2}
\end{equation}
where $C_{a},C_{p},C_{s}$ denote the sets of connections that related to the three types of potentials, respectively. 
Importantly, since anatomical priors are spatial-variant for medical scans, we define the prior potentials $\phi_{p}(l_{T}|T, A, L_{A})$ to only exist between voxels of the corresponding regions on $L_{T}$ and $L_{A}$ for retaining structured predictions. 
Moreover, considering the uniqueness of medical imaging from nature images that different anatomies in long range can have similar intensities, the smoothness potentials $\phi_{s}(l_{T}|T)$ are also restricted within neighboring voxels, which is distinct from existing graphs. 
As such, the CRF is locally-connected, and we show it enables the CAM to be compact, and can be formulated as fully convolutional. 

\textbf{Appearance Potential} 
$\phi_{a}(l_{T}|T)$ models the cost of label assignments by considering the scan appearance, \textit{i.e.}, shape and texture. 
In our model, we utilize the ConvNet segmenter to capture the potential by taking the inverse of the output likelihood over the label. 
Since the potential is derived from image patches without any constraints, it can be noisy and vulnerable to appearance variations. 

\textbf{Prior Potential}
$\phi_{p}(l_{T}|T, A, L_{A})$ measures the cost of assigning the label $y$ conditioned on atlas labels as priors.  
Specifically, the potential encourages the label consistency between voxels of corresponding locations on the target scan and the atlas scan with similar appearances.
Let $p$ and $q$ be two voxels in the target and atlas scan, the potential of their connection in our model have the form: 
\begin{equation} 
  \begin{array}{l}
  \phi_{p}(l_{T, p}|X,A,L_{A}) = \mu(l_{T, p}, A_{q}) \cdot \omega_{p} \cdot \exp(-\frac{|T_{p} - A_{q}|^2}{2{\theta_{p}}^2}), 
  \end{array}
\label{eq3}
\end{equation}
where parameter $\theta_{p}$ is a scalar that controls the degree of label propagation with a given similarity of the two voxels. 
$\mu(.,.)$ is a label compatibility function that assigns penalties for assigning different labels, which is implemented as a 2-D matrix by following \cite{zheng2015conditional}. 
$\omega_p \in \mathbb{R}^{H \times W \times T}$ is the spatial weights for the potential, such that the model learns the impact of the prior potential to predictions with high resolution. 
We define that prior potential exists between a target voxel and $S^3$ atlas voxels within the spatially corresponding region, where $S$ is the graph connectivity.   
In order to account for possible misalignment of anatomies, we enlarge the coverage of the potential by adding dilations to the connections while keeping their number fixed.
Thus, given a dilation rate of $r$, the effective field of a target voxel is $(rS -1)^3$. 

Because of the locality of connections, we are able to compute the mean field inference (summarized in \textit{Appendix \ref{apd_mf}}) involving the prior potential term without relying on an extra permutohedral lattice approximation (\cite{adams2010fast}) of the graph.
Specifically, the passing message of the prior potentials within the clique of a target voxel $i$ is equivalent to a dilated convolution (\cite{yu2015multi}):
\begin{equation} 
  \begin{array}{l}
  \sum_{C_{p}} \omega_p \cdot \exp(-\frac{|T_i-A_j|}{2\theta_p^2}) \cdot Q_j(l_T) = \\
  ~~~~~~~~~~~~~~~~~~
  \sum_{d \leq S^3}K[i, r \cdot d] \cdot Q_{i + r \cdot d}(l_T), 
  \end{array}
\label{eq4}
\end{equation}
where $Q(l_T)$ denotes an intermediate estimation of labeling distribution during mean field inference, with its subscript denotes a voxel index.
$K[i, r \cdot d] := \omega_p \cdot \exp(-\frac{|T_i-A_{i+r \cdot d}|}{2\theta_p^2})$ is the dilated kernel, with $r$ representing the dilation rate of the connections. 
The optimal parameters of the potential, \textit{i.e.} $\phi(.,.),\omega_p,\theta_p$, are all learnable from back-propagation. 

\textbf{Smoothness Potential}
$\phi_{s}(y|X)$ encourages the smoothness of the label map by modeling the cost of two voxels of similar appearance having different labels. 
The potential is introduced since both appearance and prior potentials lack the constraints for label smoothness, which can lead to poor delineations and noisy outputs. 
Again, we define the potential only exist between a target voxel and its surrounding voxels within a $S^3$-connectivity neighborhood, since different anatomies in long range can have similar appearances within medical imaging. 
Similar with the prior potential, we characterize the smoothness potential of a target voxel $i$ with its neighboring voxel $j$ by following \cite{krahenbuhl2011efficient}: 
\begin{equation} 
  \begin{array}{l}
  \phi_{s}(l_{T,i}|T) = \mu(l_{T,i}, l_{T,j}) \cdot \omega_{s} \cdot \exp(-\frac{|T_{i} - T_{j}|^2}{2{\theta_{s}}^2}),
  \end{array}
\label{eq5}
\end{equation}
where $\theta_s$ is a scalar parameter controlling the degree of label propagation given two voxel similarities, $\mu(., .)$ is the same label compatibility function as in the prior potential, and $w_s \in \mathbb{R}^{K}$ denotes a class-wise potential weight parameter.

As such, similar with the prior potential, we are able to convert its passing message as a regular convolution by utilizing this local connectivity: 
\begin{equation} 
  \begin{array}{l}
  \sum_{C_{s}} \omega_s \cdot \exp(-\frac{|T_i-T_j|}{2\theta_s^2}) \cdot Q_j(l_T) = \\
  ~~~~~~~~~~~~~~~~~~
  \sum_{d \leq S^3}K[i, d] \cdot Q_{i + d}(l_T), 
  \end{array}
\label{eq6}
\end{equation} 
where $Q_j(l_T)$ denotes a labeling estimation of a neighbouring voxel $j$, and $K[i, d] := \omega_s \cdot \exp(-\frac{|T_i-T_j|}{2\theta_s^2})$ is the convolution kernel.

\subsection{Training Strategies}
We train our models in two stages. 
In the first stage, we train the backbone ConvNets for effective appearance potentials. 
Then we train the whole model with the CAM plugged in for the optimal prior fusion performance. 
We utilize an Adam optimizer for both stages, with learning rates of 5e-4 for all parameters, except for the spatial weight $w_p$ of the prior potential having a learning rate of 1e-2.
Early stopping is employed based on the validation set to avoid over-fitting. 
For both stages of the training, we adopt a generalized multi-class dice loss (\cite{sudre2017generalised}) that is evenly weighted for all categories within training patches. 
Standard pre-processing steps are carried out, including affine spatial normalization to the MNI space and intensity standardization.
Random noises are added for augmentation. 
\section{Experiments}
We show the effectiveness of our method with two brain  parcellation datasets, and the flexibilty with three different ConvNets as backbone. 
We first compare the accuracy of our method with state-of-the-art results from ConvNet-based methods.
We then validate the robustness of our method to scan variations by introducing unforeseen pathologies to the testing scans.
We further verify the robustness to scan-level variations by cross-dataset training: training and testing models on   different datasets, where scans are collected with distinct setups.
Finally, we also perform ablation tests on different CRF  potentials to understand the impact of those factors to the performance.
In our implementation, for the prior potentials, we set the neighborhood size $S=5$ and scale factor $r=2$, which results in a local region of spatial size 9 for each voxel to be conditioned on. 
Moreover, for the smoothness potential, we set the neighborhood size to be $S = 5$ without dilation. 

\subsection{Datasets and Evaluation Metrics}
\textbf{Datasets} 
We conduct experiments on IBSR\footnote{https://www.nitrc.org/projects/ibsr} (International Brain Segmentation Repository) and MICCAI2012 (Multi-Atlas Labeling Challenge \cite{landman2012miccai}) , which are two anatomical segmentation benchmarks of T1-weighted MR brain scans. 
IBSR consists of 18 scans with annotations of 32 structures.
We randomly split the dataset into training/validation/testing subsets of 10/2/6 scans, and report the average results from three-fold tests. 
MICCAI2012 consists of 35 scan-label pairs with annotations of 134 structures, and has been pre-split into 15/30 scans for training/testing as released. 
We follow the data split, and randomly select 2 scans from the training subset for validation. 

\textbf{Evaluation Metrics}
We carry out comprehensive evaluation of accuracy using three metrics: Dice Score Coefficient (DSC), mean surface distance (MSD), and 95$\%$ Hausdorff distance (95HD). 
While DSC measures the overall overlaps between predictions and ground-truth, MSD and 95HD can quantify the amount of abnormalities leading to structure inconsistencies, such that enable us to study the effectiveness of anatomical constraints.

\begin{table}[h]
\centering
\caption{Comparison of accuracy by DSC, MSD, and 95HD as metrics. All values are illustrated in the format of mean$\pm$std. Our models significantly boost backbone ConvNets, and outperforms the state-of-the-art ConvNet methods. Bold marks best the performance among our models. Note Ganaye (20) extends Ganaye (0) with 20 unlabeled scans for model training.}
\scalebox{0.9}{
\begin{tabular}{l c c c c}
\hline
 \multicolumn{5}{c}{IBSR}\\
Method &  DSC (\%) & MSD (mm) & 95HD (mm) & Runtime (s) \\
\hline
Ganaye (0) & 83.3 $\pm$ 10.0 & 0.78 $\pm$ 0.31 & - & - \\
Ganaye (20) & 83.3 $\pm$ 10.0 & 0.77 $\pm$ 0.31 & - & - \\
CompareNet & 84.5 $\pm$ 1.8 & - & - & 42.0 \\
\hline
UNet & 82.15 $\pm$ 2.54 & 1.29 $\pm$ 1.02 & 4.20 $\pm$ 2.82 & 5.8 \\
VNet &  80.85 $\pm$ 2.47  &  0.95 $\pm$ 0.19  & 3.31 $\pm$ 1.06  & 4.6 \\
HCNet & 81.54 $\pm$ 0.84 & 1.00 $\pm$ 0.29  & 3.47 $\pm$ 0.92 & 10.8 \\
\hline
UNet-CAM (ours) & \textbf{85.67 $\pm$ 1.39} & 0.77 $\pm$ 0.15 &2.16 $\pm$ 0.35 & 12.7 \\
VNet-CAM (ours) & 85.23 $\pm$ 1.20  & \textbf{0.72 $\pm$ 0.22} & \textbf{2.05 $\pm$ 0.47} & 12.5 \\
HCNet-CAM (ours) & 83.11 $\pm$ 0.98 & 0.78 $\pm$ 0.13 & 2.90 $\pm$ 0.77 & 29.6 \\
\hline
\hline
\end{tabular}
}
\scalebox{0.9}{
\begin{tabular}{l c c c c}
\multicolumn{5}{c}{MICCAI2012}\\
Method &  DSC (\%) & MSD (mm) & 95HD (mm) & Runtime (s) \\ 
\hline
Ganaye (0) & 73.0 $\pm$ 10.0 & 1.13 $\pm$ 0.35 & - & - \\
Ganaye (20) & 73.3 $\pm$ 10.0 & 1.08 $\pm$ 0.33 & - & - \\
CompareNet & 74.6 $\pm$ 6.4 & - & - & 78.0 \\
\hline
UNet & 74.15 $\pm$ 3.57 & 1.92 $\pm$ 0.96 & 5.48 $\pm$ 3.04 & 7.2 \\
VNet & 72.14 $\pm$ 3.65 & 2.13 $\pm$ 0.94 & 5.96 $\pm$ 3.02 & 7.1 \\
HCNet & 71.08 $\pm$ 4.49 & 2.50 $\pm$ 1.65 & 6.53 $\pm$ 3.56 & 15.1 \\
\hline
UNet-CAM (ours) & \textbf{76.34 $\pm$ 3.38} & \textbf{0.91 $\pm$ 0.12} & \textbf{5.29 $\pm$ 3.79} & 30.8 \\
VNet-CAM (ours) & 75.89 $\pm$ 3.05 & 1.03 $\pm$ 0.33 & 5.47 $\pm$ 3.35 & 30.7 \\
HCNet-CAM (ours) & 73.22 $\pm$ 4.78 & 1.81 $\pm$ 0.70 & 6.02 $\pm$ 2.70 & 66.1 \\
\hline
\end{tabular}
}
\label{tab1}
\end{table}

\subsection{Comparison of Accuracy}
\label{sect_acc} 
We evaluate the accuracy of our method on the original benchmarks.
We compare with the reported results from \cite{ganaye2018semi} (Ganaye) and \cite{liang2019comparenet} (CompareNet), which are two ConvNets employing the loss-based prior fusion, and have achieved state-of-the-art accuracy on both benchmarks. 
We employ three general-purpose ConvNet segmenters as backbones of our method, and also train them separately to demonstrate the boosting effect of CAM.

As shown in Table \ref{tab1}, our models (UNet-CAM and VNet-CAM) outperform the state-of-the-art accuracy of ConvNets for both IBSR and MICCAI2012 datasets according to all the three metrics.
Among the models, UNet-CAM has the highest DSC accuracy of 85.67\% and 76.34\%, while HCNet-CAM has the lowest ones, possibly because the backbone ConvNet is less discriminate. 
Furthermore, we compare our models (backbone-CAM) with the standalone ConvNet segmenters using the paired t-test, and observe that CAM boosts all the backbone models significantly ($p<$0.0001) by DSC on both benchmarks.
This validates the flexibility of our method for boosting various ConvNets. 
Moreover, MSD and 95HD results show our method consistently
reduces the abnormal predictions that conflict with anatomy structures, which indicates our approach of fusing atlas priors as constraints is effective.  

Table \ref{tab1} also includes the runtime per scan for different models, which is measured under a controlled environment of one single Nvidia Titan Xp GPU with a batch size of 1. 
Our method runs with $2.55\times$ and $4.33\times$ the time as its backbone ConvNet on average for the two datasets. 
The increments of runtime (11.2s/32.7s on average for IBSR/MICCAI2012) are caused by the mean field inference of the proposed CAM.

\begin{figure}[h] 
\centering
\includegraphics[width=0.6\textwidth]{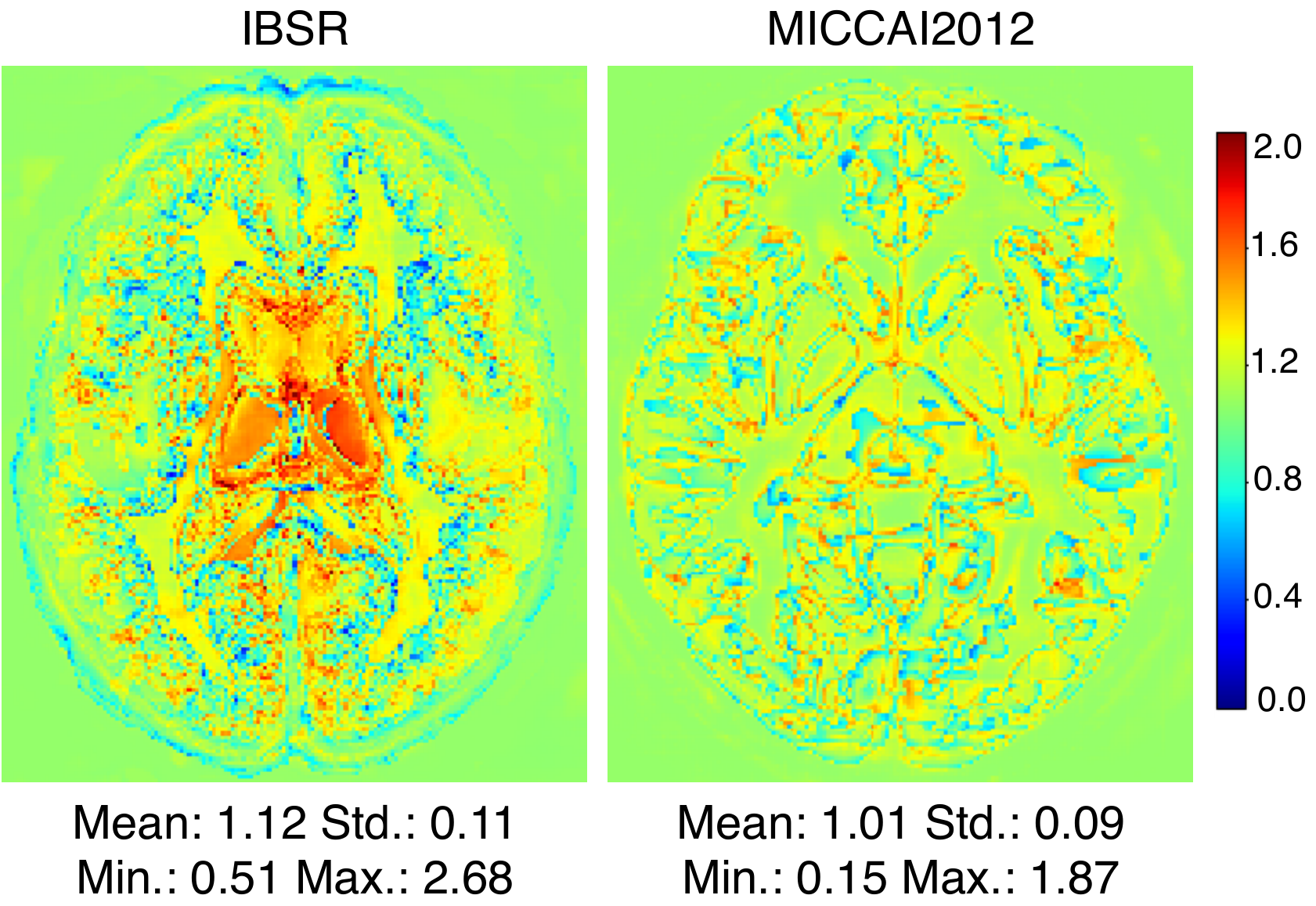}
\caption{Visualization of the fusion weights for the prior potentials that learnt from the end-to-end training. }
\label{fig_weights}
\end{figure}

We also investigate the learnt fusion weight of the atlas potential to understand its role in the predictions. 
We visualize the spatial weights of UNet-CAM as an example in Figure \ref{fig_weights} with their cutting slices. 
We observe that the weights have larger values within the anatomies, indicating that the atlas priors set a stronger constraints at the places, and weaker ones at the boundaries between adjacent anatomies. 
This further proves that the atlas priors improve the performance mainly by reducing abnormalities that conflict with the underlying anatomical structure. 
Moreover, the strong constraints for the structure also show the potential of our method for improving the model robustness against variations. 

\begin{table}[h]
\caption{Segmentation accuracy changes with the synthetic pathologies introduced. $\Delta$DSC shows the mean DSC changes, while LDD shows the mean of the local DSC decay. All values are in percentage (\%).}
\centering
\scalebox{0.9}{
\begin{tabular}{l c c c c}
\hline
\multicolumn{5}{c}{IBSR} \\
Method & DSC w/o path. & DSC w/ path. & $\Delta$DSC & LDD \\
\hline
UNet  & 82.15 & 80.19 & -1.96 & 7.43~(1$\times$) \\
UNet-CAM & 85.67 & 84.52 & \textbf{-1.15} & \textbf{4.36~(0.59$\times$)} \\
\hline
VNet  & 80.85 & 78.68 & -2.17 & 5.96~(1$\times$) \\
VNet-CAM  & 85.23 & 83.57  & \textbf{-1.66} & \textbf{3.60~(0.60$\times$)}  \\
\hline
HCNet  & 81.54 & 79.66 & -2.19 & 6.02~(1$\times$)  \\       
HCNet-CAM  & 83.11 & 82.38 &\textbf{-0.73} & \textbf{2.90~(0.48$\times$)}  \\
\hline
\hline
\end{tabular}
}
\scalebox{0.9}{
\begin{tabular}{l c c c c}
\multicolumn{5}{c}{MICCAI2012} \\
Method & DSC w/o path. & DSC w/ path. & $\Delta$DSC & LDD \\
\hline
UNet & 74.15 & 71.14 & -3.01 & 8.20~(1$\times$)\\
UNet-CAM  & 76.34 & 75.63 & \textbf{-0.71} & \textbf{3.82~(0.47$\times$)} \\
\hline
VNet  & 72.14 & 70.04 & -2.10 & 6.09~(1$\times$) \\
VNet-CAM  & 75.89 & 74.81 & \textbf{-1.08} & \textbf{2.94~(0.48$\times$)} \\
\hline
HCNet  & 71.08 & 68.33 & -2.75 & 6.91~(1$\times$) \\
HCNet-CAM  & 73.22 & 72.02 & \textbf{-1.20} & \textbf{3.17~(0.46$\times$)} \\
\hline
\end{tabular}
}
\label{tab_path}
\end{table}

\begin{figure}[h] 
\centering
\includegraphics[width=0.95\textwidth]{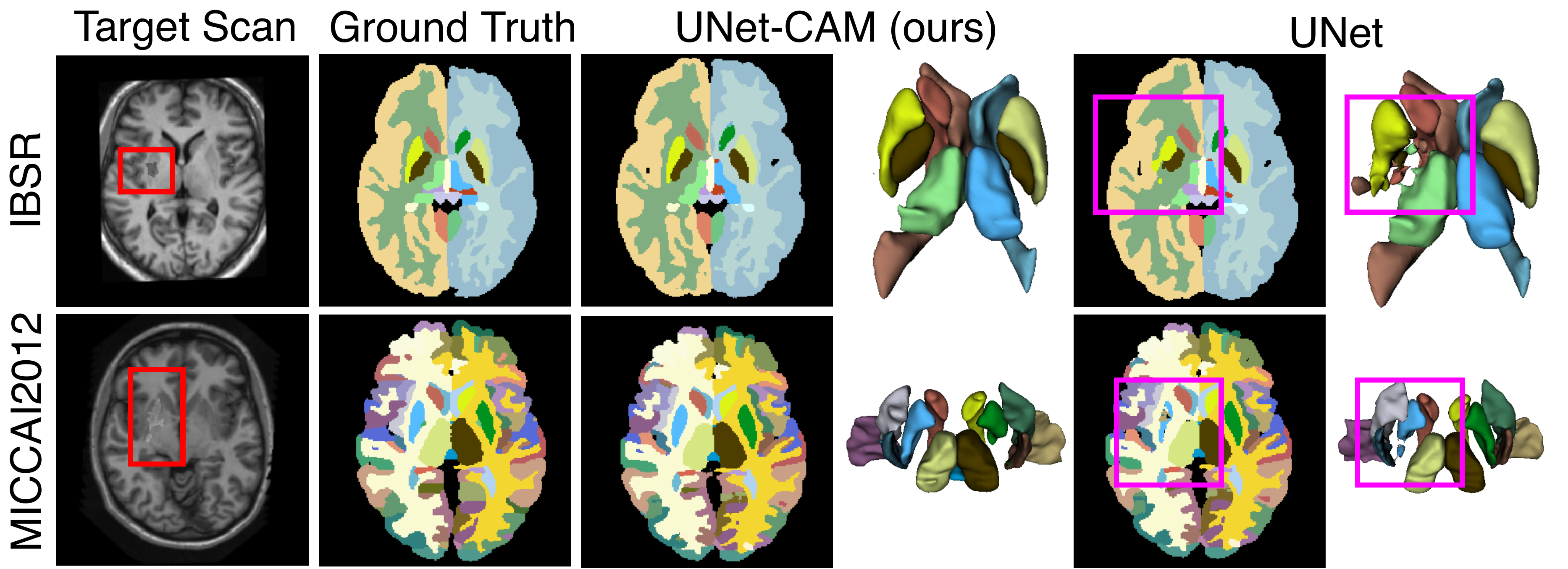}
\caption{Segmentation of UNet-CAM and UNet on two pathological scans from IBSRv2 and MICCAI testing sets, where pathological regions are delineated with red boxes. Pink boxes highlight inconsistent and noisy anatomy predictions from UNet caused by appearance variations. Meanwhile, predictions of our models are robust and accurate. 
}
\label{fig_demo}
\end{figure}
\subsection{Robustness Analysis on Scans with Pathologies}
\label{sect_path} 
In order to formally study the model robustness, we follow the method of \cite{dey2018compnet} by augmenting the testing scans of the two benchmarks with synthesized pathologies. 
To model spatial details of pathologies, $i.e.$, shapes and locations, we transfer patterns of pathologies from ATLAS (\cite{liew2018large}), which is a public brain lesion dataset with manual lesion masks.
Specifically, a lesion mask is aligned and adapted to a target scan by performing affine registrations.
To simulate the texture of pathologies, we augment intensities of pathological volumes with random noise of magnitudes ranging from 20\% to 50\% of the scan's maximum intensity.
We generate 60 augmented cases for each of the datasets from their testing data with random pathological masks and textures. 
We evaluate the models that trained on the original scans without pathologies as described in Section \ref{sect_acc}. 
Examples of the augmented scans are visualized in \textit{Supplementary Material A}. 

Table \ref{tab_path} shows the results on the augmented testing scans of different models.  
While all the models have reduced performance with the presence of pathologies, our models (backbone-CAM) obtain less accuracy changes ($\Delta$DSC) compared to their backbones. 
We also define Local Dice Decay (LDD) as the mean changes of DSC averaged over classes that have voxels within the pathological regions according to pathological masks. 
This is according to the observation that mis-predictions caused by pathologies are mostly local to the image variations (\cite{dey2018compnet}), and the LDD can better reflect the impact of image variations without that being averaged out by the large amount of unaffected labels. 
As indicated by LDD results, our method consistently improves  the model robustness for all the models: the accuracy changes can be reduced to up to 0.48$\times$ and 0.46$\times$ on IBSR and MICCAI2012 datasets, respectively. 
The robustness gain can be explained as the atlas priors effectively constraint the predictions of anatomies, such that the labeling error of large structural divergences can be reduced. 

We also visualize the results from UNet-CAM and a standalone UNet on pathological images in Figure \ref{fig_demo}. 
We highlight pathological regions and related predictions with red and pink boxes, separately.
It clearly shows that structures of anatomies are better retained with our method (UNet-CAM), while the baseline method (UNet) has fragmented anatomies under the unforeseen noises.  
More examples of segmentation on pathological scans are visualized in \textit{Supplementary Material B}. 

\subsection{Robustness Analysis on Cross-Dataset Training}
We further challenge the robustness of our method with scan-level variations by cross-dataset training: we train and evaluate models with scans from two separate datasets.
Since IBSR and MICCAI2012 are collected with different scanners and scanning parameters, the experiment simulates the possible situation that algorithms can meet in clinical applications. 
Specifically, we follow the same dataset splits described in Section \ref{sect_acc}.  
we apply the trained models based on one of the two datasets to the testing scans from the other dataset.
Note that all the  atlas  priors  applied  here  are  also  generated with the training scans, in order to simulate the gap between prior knowledge and the target scans.
Moreover, since the two datasets are annotated for different purposes, we evaluate on the only 15 anatomies that have been labeled with the same protocol in both datasets.
A list of the overlapping anatomies is as attached in  \textit{Supplementary Material C}. 
We  compare with ConvNets without atlas constraints to evaluate the effectiveness of our method. 

\begin{table}[h]
\caption{Cross-dataset Segmentation results. Note the evaluation is based on the 15 anatomies that commonly annotated in both datasets. }
\centering
\scalebox{0.9}{
\begin{tabular}{l c c c | c}
\hline
\multicolumn{5}{c}{Training on IBSR and testing on MICCAI2012} \\
\hline
Method & DSC (\%) & MSD (mm) & 95HD (mm) & $\Delta${DSC} (\%) \\
\hline
UNet  & 82.42$\pm$3.79 & 3.74$\pm$2.47 & 11.70$\pm$9.35 & -4.68~(1$\times$)   \\
UNet-CAM  & \textbf{85.28$\pm$2.21} & \textbf{1.32$\pm$1.02} & \textbf{3.94$\pm$3.01} & \textbf{-2.21~(0.47$\times$)} \\
\hline
VNet  & 82.48$\pm$3.34 & 2.67$\pm$3.50 & 8.47$\pm$12.63 & -5.91~(1$\times$)  \\
VNet-CAM  & \textbf{85.15$\pm$2.81} & \textbf{2.66$\pm$3.08} & \textbf{4.69$\pm$3.89} & \textbf{-2.38~(0.40$\times$)}\\
\hline
HCNet  & 82.46$\pm$4.01 & 3.39$\pm$2.95  & 9.14$\pm$8.50 & -3.43~(1$\times$)  \\
HCNet-CAM  & \textbf{83.83$\pm$2.80} & \textbf{1.45$\pm$0.97} & \textbf{4.34$\pm$4.19} & \textbf{-1.82~(0.53$\times$)} \\
\hline
\hline
\end{tabular}
}
\scalebox{0.9}{
\begin{tabular}{l c c c | c}
\multicolumn{5}{c}{Training on MICCAI2012 and testing on IBSR} \\
\hline
Method & DSC (\%) & MSD (mm) & 95HD (mm) & $\Delta${DSC} (\%) \\
\hline
UNet  & 80.53$\pm$6.92 & 5.35$\pm$4.49 & 15.14$\pm$14.07 & -7.74~(1$\times$)  \\
UNet-CAM  & \textbf{85.57$\pm$4.25}  & \textbf{1.79$\pm$1.54} & \textbf{4.22$\pm$3.12} & \textbf{-3.84~(0.50$\times$)}\\
\hline
VNet  & 80.49$\pm$6.60 & 2.83$\pm$1.57 & 9.48$\pm$7.61 & -7.44~(1$\times$) \\
VNet-CAM  & \textbf{82.74$\pm$2.78} & \textbf{1.15$\pm$0.41} & \textbf{3.06$\pm$1.83} & \textbf{-3.77~(0.51$\times$)}\\
\hline
HCNet  & 80.28$\pm$7.18 & 4.60$\pm$3.67 & 11.23$\pm$8.78 & -6.93~(1$\times$)  \\
HCNet-CAM  & \textbf{82.15$\pm$1.50} & \textbf{2.30$\pm$1.04} & \textbf{5.20$\pm$4.39} & \textbf{-3.28~(0.47$\times$)} \\
\hline
\end{tabular}
}
\label{tab_cross}
\end{table}

Table \ref{tab_cross} clearly shows that our method (backbone-CAM) achieves significantly ($p<$0.0001) higher accuracy according to DSC than ConvNets without atlas constraints. 
The MSD and 95HD results are even more obvious, indicating that our results have much less anatomical inconsistencies with the boost of atlas constraints. 
Moreover, to better compare the robustness, we compute the DSC changes ($\Delta$DSC) between the results on the testing scans from the same dataset as training, and the ones from the other dataset. 
The results show that our method can better retain accuracy, reducing the $\Delta$DSC from 0.40$\times$ to 0.53$\times$ as the ConvNets without atlas constraints.  

\begin{table}[h]
\caption{Ablation results for CRF potentials. Accuracy is measure with DSC, while robustness is measured with LDD on pathology-augmented scans.}
\centering
\scalebox{0.9}{
\begin{tabular}{l c c | c c}
\hline
& \multicolumn{2}{c}{IBSR} & \multicolumn{2}{c}{MICCAI2012}\\
 & Accuracy Test & Robustness Test & Accuracy Test & Robustness Test \\ 
Method & DSC (\%) & LDD (\%) & DSC (\%) & LDD (\%)\\
\hline
UNet (baseline)  & 82.15  & 7.43  & 74.15 & 8.20 \\
\hline 
UNet-SP  & 83.17 (+1.02) & 7.65 (1.03$\times$) & 74.95 (+0.80) & 8.22 (1.00$\times$)  \\
UNet-PP  &  85.32 (+3.17) & 4.31 (0.58$\times$)  & 75.98 (+1.83) & 3.90 (0.48$\times$) \\
UNet-CAM (ours)  &  85.67 (+3.52)  & 4.36 (0.59$\times$) & 76.34 (+2.19) & 3.82 (0.47$\times$)\\
\hline 
UNet-CAM$\setminus$SP  & 85.12 (+2.97) & 5.19 ($0.70\times$) & 75.60 (+1.45) & 4.23 ($0.52\times$)  \\
UNet-CAM$\setminus$PP  &  73.60 (-8.55) & 9.06 ($1.22\times$) & 63.23 (-10.92) & 7.95 ($0.97\times$) \\
UNet-CAM$\setminus$(SP+PP)  & 67.95 (-14.20) & 8.95 ($1.20\times$) & 57.60 (-16.55) & 8.12 ($0.99\times$) \\
\hline
UNet-CAM-SEPARATE  &  84.45 (+2.30)  & 5.70 ($0.77\times$) & 74.99 (+0.84) & 7.87 ($0.96\times$) \\
\hline 
\end{tabular}
}
\label{tab_potentials}
\end{table}
\subsection{Ablation Test}

We further perform ablation tests to study the impact of the three types of potential, \textit{i.e.} appearance potential, prior potential, and smoothness potential, to the accuracy and robustness of our method. 
We first train the following models separately: \textit{(i)} standalone UNet (UNet) as the baseline, \textit{(ii)} UNet with prior potentials only (UNet-PP), \textit{(iii)} UNet with smoothness potentials only (UNet-SP), and \textit{(iv)} UNet with both potentials (UNet-CAM). 
Then, to investigate the role of each potential within the CRF, we derive the following models by removing certain potentials
from the trained UNet-CAM models during inference: \textit{(i)} standalone UNet (UNet-CAM$\setminus$(SP+PP)), \textit{(ii)} UNet-CAM without prior potentials (UNet-CAM$\setminus$PP), and \textit{(iii)} UNet-CAM without smoothness potentials (UNet-CAM$\setminus$SP).
Moreover, to investigate the impact of the end-to-end training, we evaluate the UNet-CAM by optimizing fusion parameters separately from the backbone ConvNets, denoted as UNet-CAM-SEPARATE.
For the aforementioned models, we measure both the accuracy on the original benchmarks, and also the robustness by reusing the pathology-augmented scans as described in Section \ref{sect_path}. 

Table \ref{tab_potentials} shows the results. 
By comparing among UNet-SP, UNet-PP, and UNet-CAM, we observe the prior potential (UNet-PP) contributes more to both accuracy (DSC of +3.17/+1.83 for IBSR/MICCAI2012) and robustness (LDD of 0.58$\times$/0.48$\times$ for IBSR/MICCAI2012) than the smoothness potential (UNet-SP).
Moreover, the smoothness potential (UNet-SP) does not benefit the model robustness (LDD of 1.03$\times$/1.00$\times$ for IBSR/MICCAI2012). 
Moreover, the combination of both potential (UNet-CAM) leads to the best accuracy, since the two types of constraints are complementary for anatomical-consistency and smoothness. 
By comparing among models with connection-removal during inference (UNet-CAM$\setminus$PP, UNet-CAM$\setminus$SP, and UNet-CAM$\setminus$(SP+PP)), we observe significant decay in accuracy and robustness for all the models than the integrated model (UNet-CAM).  
This indicates that the optimal performance of our method (UNet-CAM) is achieved by the fully fusion of different potentials. 
Particularly, we see the removal of the prior potential (UNet-CAM$\setminus$PP and UNet-CAM$\setminus$(SP+PP)) results in largest under-performance of accuracy and robustness, which also reflects its key role for the performance of UNet-CAM. 
By comparing between UNet-CAM and UNet-CAM-SEPARATE, we show the end-to-end training of the whole model is important for the gains of accuracy (IBSR: DSC of +3.52 \textit{v.s.} +2.30; MICCAI: DSC of +2.19 \textit{v.s.} +0.84) and robustness (IBSR: LDD of 0.59$\times$ \textit{v.s.} 0.77$\times$; MICCAI: LD of 0.47$\times$ \textit{v.s.} 0.96$\times$). 
It can be explained as the backbone ConvNet is aware of the CRF during training, such that the parameters of both parts can be co-optimized for the optimal performance. 

\section{Conclusion}
In this paper, we aim to alleviate anatomical invariances among medical scans as atlas-based constraints for accurate and robust segmentation. 
We design a Constraint Adoption Module (CAM) for the propose, which has two key features.
First, CAM formulates a CRF over an atlas volume and a target volume to effectively propagate the prior labels for the knowledge fusion. 
Second, CAM is locally connected to retain prior anatomical structures, which also enables it to be formulated as convolutions without permutohedral lattice approximations. 
With the compact and convolutional CAM, we can optimize the ConvNets with fusion parameters from end-to-end for the optimal performance. 
Extensive experiments on two brain segmentation benchmarks show the atlas constraint can significantly boost the accuracy of segmentation by largely reducing anatomical inconsistencies, as well as increase the robustness.
Moreover, our experiments are based on three ConvNets of two different architectures as backbone to verify its flexibility. 
As such, our method can be easily adopted by plugging in CAM to the existing ConvNets, and it can be appealing for clinical applications where scans are collected under various setups or have different qualities. 


\begin{thebibliography}{32}
\providecommand{\natexlab}[1]{#1}
\providecommand{\url}[1]{\texttt{#1}}
\expandafter\ifx\csname urlstyle\endcsname\relax
  \providecommand{\doi}[1]{doi: #1}\else
  \providecommand{\doi}{doi: \begingroup \urlstyle{rm}\Url}\fi

\bibitem[Adams et~al.(2010)Adams, Baek, and Davis]{adams2010fast}
Andrew Adams, Jongmin Baek, and Myers~Abraham Davis.
\newblock Fast high-dimensional filtering using the permutohedral lattice.
\newblock In \emph{Computer Graphics Forum}, volume~29, pages 753--762. Wiley
  Online Library, 2010.

\bibitem[Arnab et~al.(2016)Arnab, Jayasumana, Zheng, and Torr]{arnab2016higher}
Anurag Arnab, Sadeep Jayasumana, Shuai Zheng, and Philip~HS Torr.
\newblock Higher order conditional random fields in deep neural networks.
\newblock In \emph{European Conference on Computer Vision}, pages 524--540.
  Springer, 2016.

\bibitem[Avants et~al.(2008)Avants, Epstein, Grossman, and
  Gee]{avants2008symmetric}
Brian~B Avants, Charles~L Epstein, Murray Grossman, and James~C Gee.
\newblock Symmetric diffeomorphic image registration with cross-correlation:
  evaluating automated labeling of elderly and neurodegenerative brain.
\newblock \emph{Medical image analysis}, 12\penalty0 (1):\penalty0 26--41,
  2008.

\bibitem[BenTaieb and Hamarneh(2016)]{bentaieb2016topology}
A{\"\i}cha BenTaieb and Ghassan Hamarneh.
\newblock Topology aware fully convolutional networks for histology gland
  segmentation.
\newblock In \emph{International Conference on Medical Image Computing and
  Computer-Assisted Intervention}, pages 460--468. Springer, 2016.

\bibitem[Blaiotta et~al.(2018)Blaiotta, Freund, Cardoso, and
  Ashburner]{blaiotta2018generative}
Claudia Blaiotta, Patrick Freund, M~Jorge Cardoso, and John Ashburner.
\newblock Generative diffeomorphic modelling of large mri data sets for
  probabilistic template construction.
\newblock \emph{NeuroImage}, 166:\penalty0 117--134, 2018.

\bibitem[Chen et~al.(2017)Chen, Qi, Yu, Dou, Qin, and Heng]{chen2017dcan}
Hao Chen, Xiaojuan Qi, Lequan Yu, Qi~Dou, Jing Qin, and Pheng-Ann Heng.
\newblock Dcan: Deep contour-aware networks for object instance segmentation
  from histology images.
\newblock \emph{Medical image analysis}, 36:\penalty0 135--146, 2017.

\bibitem[Dalca et~al.(2018)Dalca, Guttag, and Sabuncu]{dalca2018anatomical}
Adrian~V Dalca, John Guttag, and Mert~R Sabuncu.
\newblock Anatomical priors in convolutional networks for unsupervised
  biomedical segmentation.
\newblock In \emph{Proceedings of the IEEE Conference on Computer Vision and
  Pattern Recognition}, pages 9290--9299, 2018.

\bibitem[Dey and Hong(2018)]{dey2018compnet}
Raunak Dey and Yi~Hong.
\newblock Compnet: Complementary segmentation network for brain mri extraction.
\newblock In \emph{International Conference on Medical Image Computing and
  Computer-Assisted Intervention}, pages 628--636. Springer, 2018.

\bibitem[Fischl et~al.(2002)Fischl, Salat, Busa, Albert, Dieterich, Haselgrove,
  Van Der~Kouwe, Killiany, Kennedy, Klaveness, et~al.]{fischl2002whole}
Bruce Fischl, David~H Salat, Evelina Busa, Marilyn Albert, Megan Dieterich,
  Christian Haselgrove, Andre Van Der~Kouwe, Ron Killiany, David Kennedy, Shuna
  Klaveness, et~al.
\newblock Whole brain segmentation: automated labeling of neuroanatomical
  structures in the human brain.
\newblock \emph{Neuron}, 33\penalty0 (3):\penalty0 341--355, 2002.

\bibitem[Fonov et~al.(2011)Fonov, Evans, Botteron, Almli, McKinstry, Collins,
  Group, et~al.]{fonov2011unbiased}
Vladimir Fonov, Alan~C Evans, Kelly Botteron, C~Robert Almli, Robert~C
  McKinstry, D~Louis Collins, Brain Development~Cooperative Group, et~al.
\newblock Unbiased average age-appropriate atlases for pediatric studies.
\newblock \emph{Neuroimage}, 54\penalty0 (1):\penalty0 313--327, 2011.

\bibitem[Fu et~al.(2016)Fu, Xu, Lin, Wong, and Liu]{fu2016deepvessel}
Huazhu Fu, Yanwu Xu, Stephen Lin, Damon Wing~Kee Wong, and Jiang Liu.
\newblock Deepvessel: Retinal vessel segmentation via deep learning and
  conditional random field.
\newblock In \emph{International conference on medical image computing and
  computer-assisted intervention}, pages 132--139. Springer, 2016.

\bibitem[Ganaye et~al.(2018)Ganaye, Sdika, and Benoit-Cattin]{ganaye2018semi}
Pierre-Antoine Ganaye, Micha{\"e}l Sdika, and Hugues Benoit-Cattin.
\newblock Semi-supervised learning for segmentation under semantic constraint.
\newblock In \emph{International Conference on Medical Image Computing and
  Computer-Assisted Intervention}, pages 595--602. Springer, 2018.

\bibitem[Gao et~al.(2016)Gao, Xu, Lu, Wu, Nogues, Summers, and
  Mollura]{gao2016segmentation}
Mingchen Gao, Ziyue Xu, Le~Lu, Aaron Wu, Isabella Nogues, Ronald~M Summers, and
  Daniel~J Mollura.
\newblock Segmentation label propagation using deep convolutional neural
  networks and dense conditional random field.
\newblock In \emph{2016 IEEE 13th International Symposium on Biomedical Imaging
  (ISBI)}, pages 1265--1268. IEEE, 2016.

\bibitem[He et~al.(2019)He, Yang, Li, Li, Chang, and Yu]{he2019non}
Xiang He, Sibei Yang, Guanbin Li, Haofeng Li, Huiyou Chang, and Yizhou Yu.
\newblock Non-local context encoder: Robust biomedical image segmentation
  against adversarial attacks.
\newblock In \emph{Proceedings of the AAAI Conference on Artificial
  Intelligence}, volume~33, pages 8417--8424, 2019.

\bibitem[Iglesias and Sabuncu(2015)]{iglesias2015multi}
Juan~Eugenio Iglesias and Mert~R Sabuncu.
\newblock Multi-atlas segmentation of biomedical images: a survey.
\newblock \emph{Medical image analysis}, 24\penalty0 (1):\penalty0 205--219,
  2015.

\bibitem[Kervadec et~al.(2019)Kervadec, Dolz, Tang, Granger, Boykov, and
  Ayed]{kervadec2019constrained}
Hoel Kervadec, Jose Dolz, Meng Tang, Eric Granger, Yuri Boykov, and Ismail~Ben
  Ayed.
\newblock Constrained-cnn losses for weakly supervised segmentation.
\newblock \emph{Medical image analysis}, 54:\penalty0 88--99, 2019.

\bibitem[Kr{\"a}henb{\"u}hl and Koltun(2011)]{krahenbuhl2011efficient}
Philipp Kr{\"a}henb{\"u}hl and Vladlen Koltun.
\newblock Efficient inference in fully connected crfs with gaussian edge
  potentials.
\newblock In \emph{Advances in neural information processing systems}, pages
  109--117, 2011.

\bibitem[Landman and Warfield(2012)]{landman2012miccai}
B~Landman and S~Warfield.
\newblock Miccai 2012 workshop on multi-atlas labeling.
\newblock In \emph{Medical image computing and computer assisted intervention
  conference}, 2012.

\bibitem[Leung et~al.(2010)Leung, Clarkson, Bartlett, Clegg, Jack~Jr, Weiner,
  Fox, Ourselin, Initiative, et~al.]{leung2010robust}
Kelvin~K Leung, Matthew~J Clarkson, Jonathan~W Bartlett, Shona Clegg,
  Clifford~R Jack~Jr, Michael~W Weiner, Nick~C Fox, S{\'e}bastien Ourselin,
  Alzheimer's Disease~Neuroimaging Initiative, et~al.
\newblock Robust atrophy rate measurement in alzheimer's disease using
  multi-site serial mri: tissue-specific intensity normalization and parameter
  selection.
\newblock \emph{Neuroimage}, 50\penalty0 (2):\penalty0 516--523, 2010.

\bibitem[Li et~al.(2017)Li, Wang, Fidon, Ourselin, Cardoso, and
  Vercauteren]{li2017compactness}
Wenqi Li, Guotai Wang, Lucas Fidon, Sebastien Ourselin, M~Jorge Cardoso, and
  Tom Vercauteren.
\newblock On the compactness, efficiency, and representation of 3d
  convolutional networks: brain parcellation as a pretext task.
\newblock In \emph{International Conference on Information Processing in
  Medical Imaging}, pages 348--360. Springer, 2017.

\bibitem[Liang et~al.(2019)Liang, Song, Dym, Wang, and He]{liang2019comparenet}
Yuan Liang, Weinan Song, JP~Dym, Kun Wang, and Lei He.
\newblock Comparenet: Anatomical segmentation network with deep non-local label
  fusion.
\newblock In \emph{International Conference on Medical Image Computing and
  Computer-Assisted Intervention}, pages 292--300. Springer, 2019.

\bibitem[Liew et~al.(2018)Liew, Anglin, Banks, Sondag, Ito, Kim, Chan, Ito,
  Jung, Khoshab, et~al.]{liew2018large}
Sook-Lei Liew, Julia~M Anglin, Nick~W Banks, Matt Sondag, Kaori~L Ito, Hosung
  Kim, Jennifer Chan, Joyce Ito, Connie Jung, Nima Khoshab, et~al.
\newblock A large, open source dataset of stroke anatomical brain images and
  manual lesion segmentations.
\newblock \emph{Scientific data}, 5:\penalty0 180011, 2018.

\bibitem[Milletari et~al.(2016)Milletari, Navab, and Ahmadi]{milletari2016v}
Fausto Milletari, Nassir Navab, and Seyed-Ahmad Ahmadi.
\newblock V-net: Fully convolutional neural networks for volumetric medical
  image segmentation.
\newblock In \emph{2016 Fourth International Conference on 3D Vision (3DV)},
  pages 565--571. IEEE, 2016.

\bibitem[Oktay et~al.(2017)Oktay, Ferrante, Kamnitsas, Heinrich, Bai,
  Caballero, Cook, De~Marvao, Dawes, O‘Regan, et~al.]{oktay2017anatomically}
Ozan Oktay, Enzo Ferrante, Konstantinos Kamnitsas, Mattias Heinrich, Wenjia
  Bai, Jose Caballero, Stuart~A Cook, Antonio De~Marvao, Timothy Dawes,
  Declan~P O‘Regan, et~al.
\newblock Anatomically constrained neural networks (acnns): application to
  cardiac image enhancement and segmentation.
\newblock \emph{IEEE transactions on medical imaging}, 37\penalty0
  (2):\penalty0 384--395, 2017.

\bibitem[Patenaude et~al.(2011)Patenaude, Smith, Kennedy, and
  Jenkinson]{patenaude2011bayesian}
Brian Patenaude, Stephen~M Smith, David~N Kennedy, and Mark Jenkinson.
\newblock A bayesian model of shape and appearance for subcortical brain
  segmentation.
\newblock \emph{Neuroimage}, 56\penalty0 (3):\penalty0 907--922, 2011.

\bibitem[Ronneberger et~al.(2015)Ronneberger, Fischer, and
  Brox]{ronneberger2015u}
Olaf Ronneberger, Philipp Fischer, and Thomas Brox.
\newblock U-net: Convolutional networks for biomedical image segmentation.
\newblock In \emph{International Conference on Medical image computing and
  computer-assisted intervention}, pages 234--241. Springer, 2015.

\bibitem[Sabuncu et~al.(2010)Sabuncu, Yeo, Van~Leemput, Fischl, and
  Golland]{sabuncu2010generative}
Mert~R Sabuncu, BT~Thomas Yeo, Koen Van~Leemput, Bruce Fischl, and Polina
  Golland.
\newblock A generative model for image segmentation based on label fusion.
\newblock \emph{IEEE transactions on medical imaging}, 29\penalty0
  (10):\penalty0 1714--1729, 2010.

\bibitem[Sudre et~al.(2017)Sudre, Li, Vercauteren, Ourselin, and
  Cardoso]{sudre2017generalised}
Carole~H Sudre, Wenqi Li, Tom Vercauteren, Sebastien Ourselin, and M~Jorge
  Cardoso.
\newblock Generalised dice overlap as a deep learning loss function for highly
  unbalanced segmentations.
\newblock In \emph{Deep learning in medical image analysis and multimodal
  learning for clinical decision support}, pages 240--248. Springer, 2017.

\bibitem[Yu and Koltun(2015)]{yu2015multi}
Fisher Yu and Vladlen Koltun.
\newblock Multi-scale context aggregation by dilated convolutions.
\newblock \emph{arXiv preprint arXiv:1511.07122}, 2015.

\bibitem[Zhao et~al.(2016)Zhao, Wu, Song, Li, Fan, and Zhang]{zhao2016brain}
Xiaomei Zhao, Yihong Wu, Guidong Song, Zhenye Li, Yong Fan, and Yazhuo Zhang.
\newblock Brain tumor segmentation using a fully convolutional neural network
  with conditional random fields.
\newblock In \emph{International Workshop on Brainlesion: Glioma, Multiple
  Sclerosis, Stroke and Traumatic Brain Injuries}, pages 75--87. Springer,
  2016.

\bibitem[Zheng et~al.(2015)Zheng, Jayasumana, Romera-Paredes, Vineet, Su, Du,
  Huang, and Torr]{zheng2015conditional}
Shuai Zheng, Sadeep Jayasumana, Bernardino Romera-Paredes, Vibhav Vineet,
  Zhizhong Su, Dalong Du, Chang Huang, and Philip~HS Torr.
\newblock Conditional random fields as recurrent neural networks.
\newblock In \emph{Proceedings of the IEEE international conference on computer
  vision}, pages 1529--1537, 2015.

\bibitem[Zhou et~al.(2019)Zhou, Li, Bai, Wang, Chen, Han, Fishman, and
  Yuille]{zhou2019prior}
Yuyin Zhou, Zhe Li, Song Bai, Chong Wang, Xinlei Chen, Mei Han, Elliot Fishman,
  and Alan Yuille.
\newblock Prior-aware neural network for partially-supervised multi-organ
  segmentation.
\newblock \emph{arXiv preprint arXiv:1904.06346}, 2019.

\end{thebibliography}

\appendix
\section{mean field Approximation of CRF Forward Propagation}\label{apd_mf}
\begin{figure}[h]
\centering
\includegraphics[width=\textwidth]{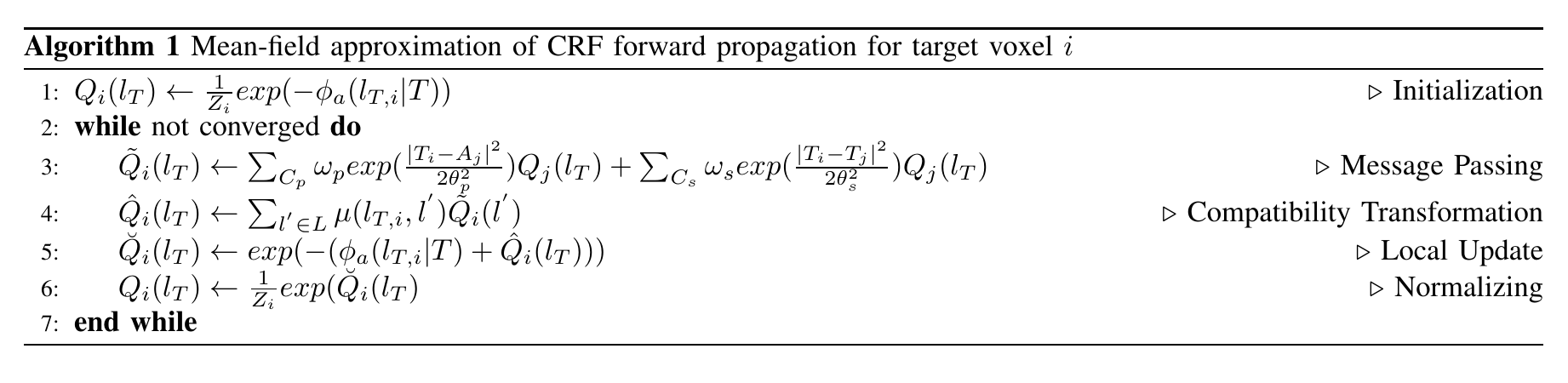}
\end{figure}

\end{document}